\documentclass[final,3p]{elsarticle}

\usepackage{lineno}
\usepackage[colorlinks=true,breaklinks=true,pdftex]{hyperref}
\usepackage{amsmath}
\usepackage{amssymb,amsfonts}

\modulolinenumbers[1]

\journal{GMP 2025}

\biboptions{authoryear}

\begin{document}

\begin{frontmatter}

\title{Geometry-Aware Face Reconstruction Under Occluded Scenes}


\author[first]{Dapeng Zhao}
\ead{mirror1775@gmail.com}
\address[first]{Zhejiang Lab, Hangzhou,China}

\begin{abstract}
    Recently, deep learning-based 3D face reconstruction methods have demonstrated promising advancements in terms of quality and efficiency. Nevertheless, these techniques face challenges in effectively handling occluded scenes and fail to capture intricate geometric facial details. Inspired by the principles of GANs and bump mapping, we have successfully addressed these issues. Our approach aims to deliver comprehensive 3D facial reconstructions, even in the presence of occlusions.While maintaining the overall shape's robustness, we introduce a mid-level shape refinement to the fundamental structure. Furthermore, we illustrate how our method adeptly extends to generate plausible details for obscured facial regions. We offer numerous examples that showcase the effectiveness of our framework in producing realistic results, where traditional methods often struggle. To substantiate the superior adaptability of our approach, we have conducted extensive experiments in the context of general 3D face reconstruction tasks, serving as concrete evidence of its regulatory prowess compared to manual occlusion removal methods.
\end{abstract}


\end{frontmatter}


\section{Introduction}
In the digital economy era, deriving a human face's shape from a single image is a cutting-edge area of research~\cite{RN823}. As individuals seek an improved quality of life, the challenge of face recognition in obstructed settings has gained significant attention, closely tied to 3D face reconstruction. Yet, many recent reconstruction methods prioritize texture accuracy, often overlooking extreme scenarios.
\\Thanks to advancements in deep learning, certain techniques can yield intricate 3D facial shapes~\cite{RN106, RN200}. However, when faces are partially occluded, these methods tend to either reconstruct occlusions indiscriminately or fail outright. In the past, researchers commonly relied on shape-from-shading (SfS) methods~\cite{RN144, RN824} to capture geometric details.
\\In most cases, general face reconstruction methods often fail to provide geometric details~\cite{RN186, RN394,RN825,RN41} or typically provide few details~\cite{RN394} to avoid reconstruction failures in extreme, challenging, unconstrained scenes. Most of these methods are based on localizing facial landmarks~\cite{RN825, RN88,RN18}
. Unlike previous work, we describe an approach designed to attain both goals: Detailed 3D face reconstruction and robustness to occluded conditions.
\\\textit{The main contributions are summarized as follows:}
\\$\bullet$\ We propose an algorithm that combines facial landmarks and the face parsing map to identify the occluded region.
\\$\bullet$\ To reconstruct the occluded area, we design a synthesis subnet that employs the predicted landmarks as guidance.
\\$\bullet$\ We propose a novel 3D face reconstruction method that can produce reasonable face geometry details under occluded scenes.
\section{RELATED ARTS}
\subsection{3D Morphable Model}
With the advancement of deep learning, significant progress has been made in 3D reconstruction, especially in the context of face reconstruction, utilizing Convolutional Neural Networks. The method proposed by Anh \textit{et al.}~\cite{RN41} offers both dense face alignment and facial model results. However, the effectiveness of these approaches is constrained by the limitations inherent in the 3D space defined by templates. 

In contrast, some recent end-to-end techniques~\cite{RN88, RN400} have achieved state-of-the-art performance in their respective domains. These methods, however, typically directly reconstruct the facial model without the need for input image identification or processing. VRN~\cite{RN88} introduces an innovative deep 3D face reconstruction approach that eliminates the necessity for precise alignment and the establishment of dense image correspondences, requiring only a single 2D facial image. VRN significantly widens the range of applications for 3D face technology. On the other hand, PRNet~\cite{RN400} has devised a straightforward method for simultaneously reconstructing the 3D facial structure. This approach incorporates a UV position map to record the 3D facial shape in UV space.
\subsection{Generative Adversarial Nets}
Under normal circumstances, we believe that Generative Adversarial Nets (GANs) are among the few tools capable of achieving a reconstruction effect in the 2D face field. After identifying and removing the occluded areas, we employ GAN to synthesize the facial image while preserving a reasonable topological structure. GAN typically comprises a generator and a discriminator, which engage in a competitive process. In the realm of 2D images, GAN stands out as one of the most effective tools for face synthesis. GAN has found successful applications in the fields of image manipulation~\cite{RN745, RN572, RN311, RN747} and image-to-image translation~\cite{RN257, RN327, RN565, RN390, RN574}.
\subsection{Facial Landmark Detection}
Traditional facial landmark detectors are more regression-based. Valstar et al.~\cite{RN842} predict landmark locations from face image patches using support vector regression (SVR). Cao et al.~\cite{RN154} use cascaded regression with pixel-difference features. Several other works~\cite{RN844,RN845,RN846} employ random regression forests to cast votes for landmark locations based on local face images. In these methods, an initial landmark position is first randomly selected, and then the facial landmarks are iteratively predicted. Therefore, the quality of the initial position is crucial.
 \section{Our Method}
 \begin{figure*}[h]
  \centering
  \includegraphics[width=0.7\textwidth]{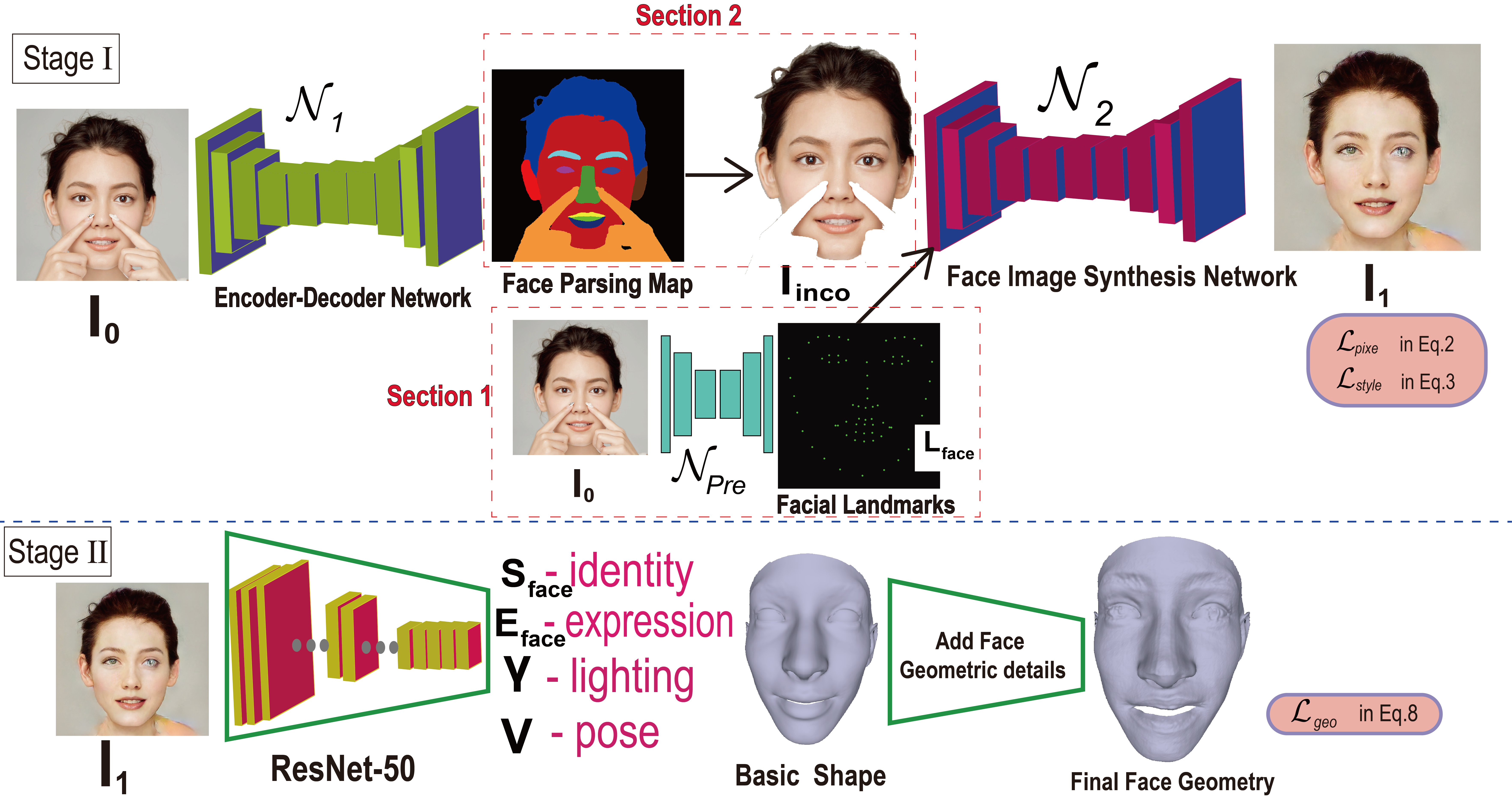}
  \caption{Our method overview.} \label{fig:overall}
\end{figure*}
As shown in Figure~\ref{fig:overall}, our method is divided into two phases: the final output of stage 1 is to remove the 2D face picture of the occlusion; the final output of stage 2 is a 3D face model with geometric details.
\subsection{Landmark Prediction Network}
Face alignment (especially accurate 
facial landmark ${{\mathbf{L}}_{\mathbf{face}}}\in {{\mathbb{R}}^{2\times 68}}$  generation) is the basis of the whole framework (Section 1 of Figure~\ref{fig:overall}). The closest approach to our method is MobileNet-V3 
model~\cite{RN740}. The network ${{\mathcal{N}}_{pre}}$ aims to generate ${{\mathbf{L}}_{\mathbf{face}}}$ from a face image ${{\mathbf{I}}_{\mathbf{0}}}:{{\mathbf{L}}_{\mathbf{face}}}={{\mathcal{N}}_{L}}({{\mathbf{I}}_{\mathbf{0}}};{{\theta }_{land}})$ , where ${{\theta }_{land}}$  denotes the parameters to be trained. Since our task is to export face features rather than identify different people, we design the loss function as follows:
\begin{equation}
  {{\mathcal{L}}_{lmk}}=\left\| {{\mathbf{L}}_{\mathbf{face}}}-{{\mathbf{L}}_{\mathbf{gt}}} \right\|_{2}^{2}
\end{equation}
where ${{\mathbf{L}}_{\mathbf{gt}}}$ is the ground truth face landmarks and ${{\left\| \cdot  \right\|}_{2}}$ is the ${{L}_{2}}$ norm.
\subsection{Face Synthesis Network}
\begin{figure}[h]
  \centering
  \includegraphics[width=0.25\textwidth]{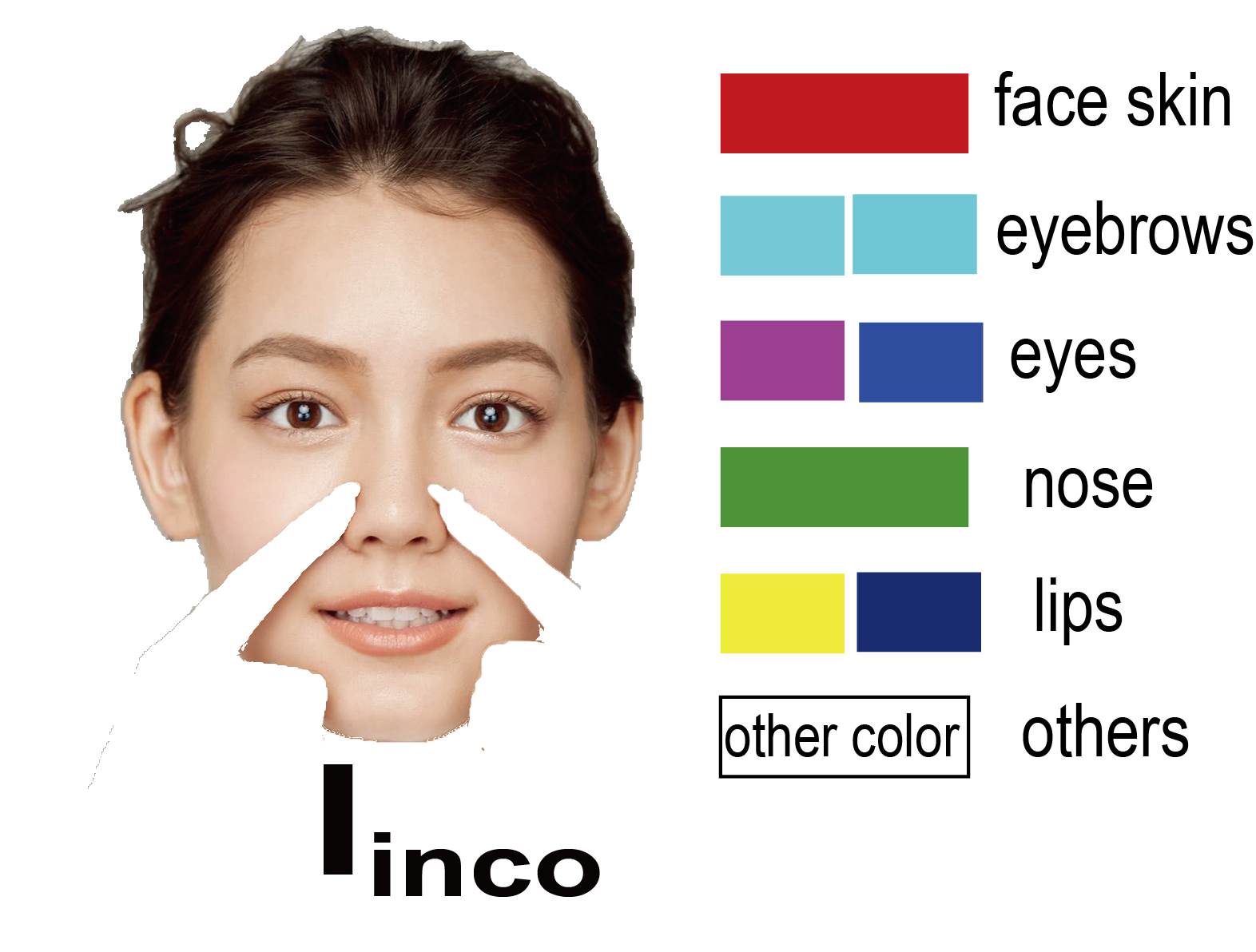}
  \caption{Face parsing network of our method.} \label{fig:faceParsing}
\end{figure}
Overall, We design the image synthesis module ${{\mathcal{N}}_{s}}$ to synthesize a 2D face that removes the occlusion region. The module includes three parts: matting operator (Section 2 of Figure~\ref{fig:overall}), generator and discriminator.
\\\textit{Matting operator.} Typically, the task of the Matting operator ${{\mathcal{N}}_{1}}$  is to remove the occluded areas ${{\mathbf{I}}_{\mathbf{m}}}$ of the input photo ${{\mathbf{I}}_{\mathbf{0}}}$ (Figure \ref{fig:faceParsing}). 
For each input face image ${{\mathbf{I}}_{\mathbf{in}}}\in {{\mathbb{R}}^{\text{H}\times \text{W}\times \text{3}}}$ , we utilized the trained 
model ${{\mathcal{N}}_{1}}$  to estimate pixel-level label classes and generate the face parsing map $\mathbf{Q}\in {{\mathbb{R}}^{\text{H}\times \text{W}\times 1}}$ .
According to the map $\mathbf{Q}$, we identify and remove the occluded 
area ${{\mathbf{I}}_{\mathbf{m}}}$ to obtain the hollowed-out photo ${{\mathbf{I}}_{\mathbf{inco}}}$.This will be the input for our next task.
\\\textit{Generator.} The generator ${{\mathcal{N}}_{2}}$ desires to generate the complete face by taking hollowed-out images ${{\mathbf{I}}_{\mathbf{inco}}}$ and landmarks $\mathbf{L}$ (${{\mathbf{L}}_{\mathbf{face}}}$  or ${{\mathbf{L}}_{\mathbf{gt}}}$). More specifically, the network includes gradually down-sampled encoding blocks. These blocks are composed of subordinate residual blocks with dilated convolutions and an attention block. The generator can be formulated as ${{\mathbf{I}}_{\mathbf{1}}}={{\mathcal{N}}_{2}}({{\mathbf{I}}_{\mathbf{inco}}},L;{{\theta }_{2}})$ , with ${{\theta }_{2}}$ the trainable parameters.
\\\textit{Discriminator.} The purpose of the discriminator is to judge whether the data distribution meets our requirements. The ambition of face synthesis is achieved when the generated results are not distinguishable from the real ones.
\\\textit{Loss Function.}We use a combination of per-pixel loss and style loss for training the face synthesis network.
\\We compute the per-pixel loss as follows:
\begin{equation}
  {{\mathcal{L}}_{pixe}}=\frac{1}{{{M}_{s}}}\left\| {{\mathbf{I}}_{\mathbf{1}}}-{{\mathbf{I}}_{\mathbf{0}}} \right\|
\end{equation}
where ${{M}_{s}}$ denotes the mask size and  $\left\| \cdot  \right\|$ stands for the ${{L}_{1}}$ norm. Notice that we use ${{M}_{s}}$ as the denominator and its role is to adjust the penalty. 
\\The style loss computes the style distance between two images as follows:
\begin{equation}
  {{\mathcal{L}}_{style}}=\sum\limits_{\text{n}}{\frac{1}{{{R}_{n}}\times {{R}_{n}}}\left\| \frac{{{B}_{n}}({{\mathbf{I}}_{\mathbf{1}}}\odot {{\mathbf{I}}_{\mathbf{m}}})-{{G}_{n}}({{\mathbf{I}}_{\mathbf{0}}}\odot {{\mathbf{I}}_{\mathbf{m}}})}{{{R}_{n}}\times {{H}_{n}}\times {{W}_{n}}} \right\|}
\end{equation}
where ${{B}_{\text{n}}}\text{(x)=}{{\varphi }_{n}}{{(x)}^{T}}{{\varphi }_{n}}(x)$  denotes the Gram Matrix corresponding to ${{\varphi }_{n}}(x)$,${{\varphi }_{n}}(\cdot )$ stands for the ${{R}_{n}}$ feature maps with the size ${{H}_{n}}\times {{W}_{n}}$ of the \textit{n}-th layer.
\\The total loss with respect to the face synthesis module:
\begin{equation}
  {{\mathcal{L}}_{final}}={{\lambda }_{pixe}}{{\mathcal{L}}_{pixe}}+{{\lambda }_{style}}{{\mathcal{L}}_{style}}
\end{equation}
here, we use ${{\lambda }_{pixe}}=1$ ,${{\lambda }_{style}}=250$  in our experiments.
\subsection{Construct the Fundamental Shapes}
We utilize the hot deep learning method in recent years to regress the 3DMM coefficients to construct the basic face shape. The primary face model is formulated as follows:
\begin{equation}
  {{\mathbf{S}}_{\mathbf{face}}}=\overline{\mathbf{S}}+\sum\limits_{i=1}^{m-1}{{{\alpha }_{i}}{{\mathbf{s}}_{\mathbf{i}}}}
\end{equation}
where $\overline{\mathbf{S}}$ denotes an average 3D face shape, $m$ denotes the number of faces participating in the weighted face 
datasets,${{\mathbf{s}}_{\mathbf{i}}}$ denotes the ith face shape and ${{\alpha }_{i}}$ denotes the face shape coefficients estimated from ${{\mathbf{I}}_{\mathbf{1}}}$.
\\Similarly, we model the expressions using the following formulation:
\begin{equation}
  {{\mathbf{E}}_{\mathbf{face}}}=\sum\limits_{j=1}^{m-1}{{{\beta }_{i}}{{\mathbf{e}}_{\mathbf{i}}}}
\end{equation}
where ${{\beta }_{i}}$ denotes the exptression coefficients and ${{\mathbf{e}}_{\mathbf{i}}}$ denotes the $i$th expression.
\\Further, we denote the viewpoint as $\mathbf{v}\text{= }\!\![\!\!\text{ }{{\mathbf{r}}^{\text{T}}}\text{,}{{\mathbf{t}}^{\text{T}}}\text{ }\!\!]\!\!$ ,where $\mathbf{r}\in {{\mathbb{R}}^{3}}$ denotes the 3D rotation, expressed by Euler angles,and $\mathbf{t}\in {{\mathbb{R}}^{3}}$ denotes a translation vector, together aligning a generic 3D face shape with the face appearing in ${{\mathbf{I}}_{\mathbf{1}}}$.
\\We adopt the Basel Face Model (BFM)~\cite{RN45}, which provides both $\overline{\mathbf{S}}$ ,${{\mathbf{s}}_{\mathbf{i}}}$ and ${{\mathbf{e}}_{\mathbf{i}}}$.We approximated the scene illumination with  Spherical Harmonics (SH)~\cite{RN317,RN772,RN239} parameterized by coefﬁcient vector $\mathbf{\gamma }\in {{\mathbb{R}}^{9}}$. Given ${{\mathbf{I}}_{\mathbf{1}}}$, we estimate $y=({{\mathbf{S}}_{\mathbf{face}}},{{\mathbf{E}}_{\mathbf{face}}},\mathbf{V},\mathbf{\gamma })$ using the recent deep 3DMM approach of 
Yu \textit{et al.}~\cite{RN239}, with their pre-trained model.
\subsection{Recovering Face Geometric Details}
Inspired by the method of image-to-image translation~\cite{RN257,RN226}, we define the displacements of the depth map as the distances through the 
pixels of ${{\mathbf{I}}_{\mathbf{1}}}$ to the 3D face surface. Generally, we define the bump map $\Phi (\mathbf{b})$  as:
\begin{equation}
    \Phi (\mathbf{b})=\left\{ \begin{aligned}
        & \phi (0)\text{other} \\ 
       & \phi ({d}'(\mathbf{b})-d(\mathbf{b}))\text{ face }\text{projects }\text{to }\mathbf{b} \\ 
      \end{aligned} \right.      
\end{equation}
where $\phi (\cdot )$ denotes an encoding function that converts the depth value to the linear range $\left[ 0,\ldots ,255 \right]$,$\mathbf{b}$ denotes the pixel coordinate $\left[ x,y \right]$  in ${{\mathbf{I}}_{\mathbf{1}}}$ , ${d}'(\mathbf{b})$  denotes the depth, which is the distance from the surface of the detailed face shape to $\mathbf{b}$ along the line of sight,$d(\mathbf{b})$  denotes the depth of the basic shape.
\\Thus, Given a bump map $\Phi $ and the depth of the basic shape, we can compute the detailed depth follows ${d}'(\mathbf{b})=d(\mathbf{b})+{{\phi }^{-1}}(\Phi (\mathbf{b}))$.
\\In order to increase geometric details and to suppress noise, we define the loss function as follows:
\begin{equation}
    {{\mathcal{L}}_{geo}}=\left\| \tilde{\Phi }-\Phi  \right\|+\left\| \frac{\partial \tilde{\Phi }}{\partial x}-\frac{\partial \Phi }{\partial x} \right\|+\left\| \frac{\partial \tilde{\Phi }}{\partial y}-\frac{\partial \Phi }{\partial y} \right\|
\end{equation}
where $\left\| \cdot  \right\|$ denotes the ${{L}_{1}}$ norm,$\tilde \Phi$  denotes the ground truth and $\frac{\partial \tilde{\Phi }}{\partial x}$ ,$\frac{\partial \tilde{\Phi }}{\partial {y}}$ denotes the 2D gradient of the bump map.
We found that by adding these last two terms of loss function and we reduce bump map noise by favoring smoother surfaces. At the same time, the final effect shows that high-frequency details are preserved.
\section{Experimental results}
\begin{figure}[h]
    \centering
    \includegraphics[width=0.45\textwidth]{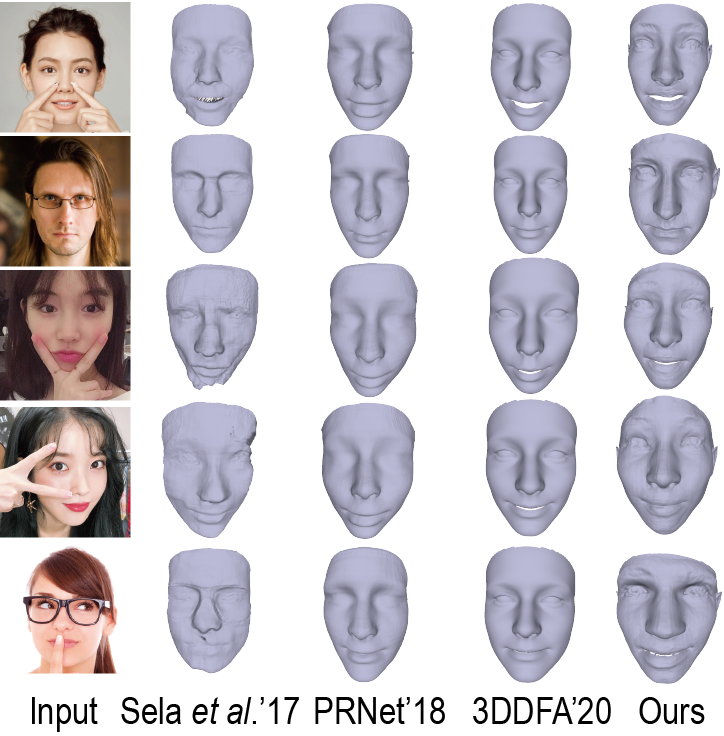}
    \caption{Comparison of qualitative results. Baseline methods from left to right: Sela \textit{et al.}, PRNet,3DDFA, and our method.} \label{fig:compare}
\end{figure}
\begin{figure}[h]
    \centering
    \includegraphics[width=0.35\textwidth]{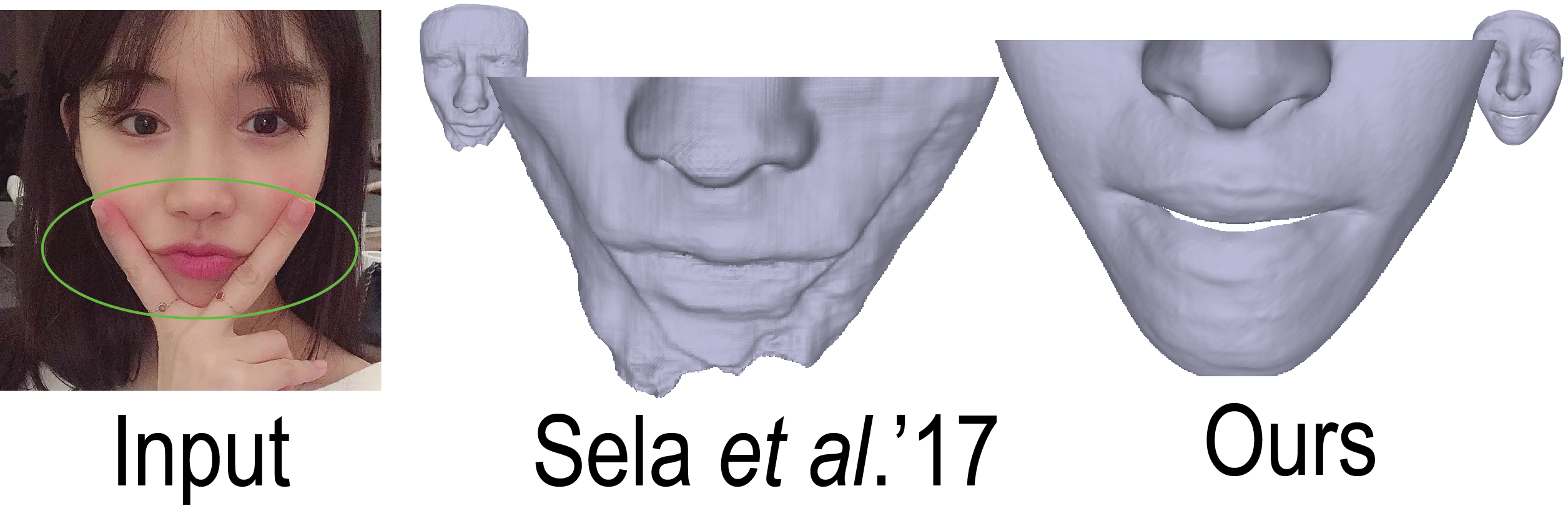}
    \caption{Comparison of local facial geometric details.} \label{fig:localcompare}
\end{figure}

Figure~\ref{fig:compare} shows our reconstruction results compared with the contemporary 
arts~\cite{RN200,RN186,RN703}. It can be clearly seen from the Figure~\ref{fig:compare} that Sela \textit{et al.}'s method shows the shape of various obstructions. This may be due to the method focusing too much on the local 
shape (Figure~\ref{fig:localcompare}). Since the other two methods are based on existing 3D face model templates, the shapes are too smooth and lack geometric details.
\section{Conclusions}
We describe an approach capable of producing detailed 3D face reconstructions from images taken in occluded conditions. The advantage of our method is that it can effectively process the facial details in places where the face is occluded. Since the previous arts are limited to the reconstruction of faces in non-occluded scenes, our method represents a leap in 3D facial details reconstruction capabilities. Unlike the recently popular end-to-end reconstruction method, our method is a two-stage face reconstruction task based on basic shapes.
\bibliography{./mybibfile.bib}
\end{document}